\documentclass{article}

\usepackage{times}
\usepackage{graphicx} 
\usepackage{subfigure} 
\usepackage{amsmath,amsthm,amsfonts}
\usepackage{natbib}

\usepackage{algorithm}
\usepackage{algorithmic}
\usepackage{graphicx}

\usepackage{hyperref}

\DeclareMathOperator{\argmin}{arg\,min}

\newcommand{\Prob}{\mathbb{P}}

\newcommand{\reals}{\mathbb{R}}

\newcommand{\y}{\mathbf{y}}
\newcommand{\z}{\mathbf{z}}
\newcommand{\w}{\mathbf{w}}
\newcommand{\x}{\mathbf{x}}
\newcommand{\rb}{\mathbf{r}}
\newcommand{\pb}{\mathbf{p}}
\newcommand{\ab}{\mathbf{a}}

\usepackage[accepted]{icml2017}

\icmltitlerunning{End-to-End SPENs}

\begin{document} 

\twocolumn[
\icmltitle{End-to-End Learning for Structured Prediction Energy Networks}




\begin{icmlauthorlist}
\icmlauthor{David Belanger}{aff1}
\icmlauthor{Bishan Yang}{aff2}
\icmlauthor{Andrew McCallum}{aff1}
\end{icmlauthorlist}

\icmlaffiliation{aff1}{University of Massachusetts, Amherst}
\icmlaffiliation{aff2}{Carnegie Mellon University}

%
%

\icmlcorrespondingauthor{David Belanger}{belanger@cs.umass.edu}

\icmlkeywords{structured prediction, deep learning, energy-based model}

\vskip 0.3in
]



\printAffiliationsAndNotice{}  


\begin{abstract} 
Structured Prediction Energy Networks (SPENs) are a simple, yet expressive family of structured prediction models~\citep{belanger2016structured}.  An energy function over candidate structured outputs is given by a deep network, and predictions are formed by gradient-based optimization. This paper presents end-to-end learning for SPENs, where the energy function is discriminatively trained by back-propagating through gradient-based prediction. In our experience, the approach is substantially more accurate than the structured SVM method of~\citet{belanger2016structured}, as it allows us to use more sophisticated non-convex energies. We provide a collection of techniques for improving the speed, accuracy, and memory requirements of end-to-end SPENs, and demonstrate the power of our method on 7-Scenes image denoising and CoNLL-2005 semantic role labeling tasks. In both, inexact minimization of non-convex SPEN energies is superior to baseline methods that use simplistic energy functions that can be minimized exactly.




\end{abstract} 

\section{Introduction}

In a variety of application domains, given an input $\x$ we seek to predict a structured output $\y$. For example, given a noisy image, we predict a clean version of it, or given a sentence we predict its semantic structure. Often, it is insufficient to employ a feed-forward predictor $\y = F(\x)$, since this may have prohibitive sample complexity, fail to model global interactions among outputs, or fail to enforce hard output constraints. Instead, it can be advantageous to define the prediction function implicitly in terms of energy minimization~\citep{lecun2006tutorial}:
\begin{equation}
\mathbf{\hat{y}} = \argmin_\y E_{\x}(\y) \label{eq:spen}, 
\end{equation}
where $E_{\x}(\cdot)$ depends on $\x$ and learned parameters. 

This approach includes factor graphs~\citep{kschischang2001factor}, e.g., conditional random fields (CRFs)~\citep{lafferty2001conditional}, and many recurrent neural networks (Sec.~\ref{sec:fac-energy}).
Output constraints can be enforced using constrained optimization. Compared to feed-forward approaches, energy-based approaches often provide better opportunities to inject prior knowledge about likely outputs and often have more parsimonious models. On the other hand, energy-based prediction requires non-trivial search in the exponentially-large space of outputs, and search techniques often need to be designed on a case-by-case basis. 

Structured prediction energy networks (SPENs)~\citep{belanger2016structured} help reduce these concerns. They can capture high-arity interactions among components of $\y$ that would lead to intractable factor graphs and provide a mechanism for automatic structure learning. This is accomplished by expressing the energy function in Eq.~\eqref{eq:spen} as a deep architecture and forming predictions by approximately optimizing $\y$ using gradient descent. 

While providing the expressivity and generality of deep networks, SPENs also maintain the useful semantics of energy functions: domain experts can design architectures to capture known properties of the data, energy functions can be combined additively, and we can perform constrained optimization over $\y$. Most importantly, SPENs provide for black-box interaction with the energy, via forward and back-propagation. This allows practitioners to explore a wide variety of models without the need to hand-design corresponding prediction methods. 

\citet{belanger2016structured} train SPENs using a structured SVM (SSVM) loss~\citep{koller2004max,tsochantaridis2004support} and achieve competitive performance on simple multi-label classification tasks. Unfortunately, we have found it difficult to extend their method to more complex domains. SSVMs are unreliable when \text{exact} energy minimization is intractable, as loss-augmented inference may fail to discover margin violations (Sec.~\ref{sec:learning-exact}). 

In response, we present end-to-end training of SPENs, where one directly back-propagates through a computation graph that unrolls gradient-based energy minimization. This does not assume that exact minimization is tractable, and instead directly optimizes the practical performance of a particular approximate minimization algorithm. End-to-end training for gradient-based prediction was introduced in~\citet{domke2012generic} and applied to deep energy models by~\citet{brakel2013training}. See Sec.~\ref{sec:unrolled} for details.

When applying end-to-end training to SPENs for problems with sophisticated output structure, we have encountered a variety of technical challenges. The core contribution of this paper is a set of general-purpose solutions for overcoming these.  Sec.~\ref{sec:simplex} alleviates the effect of vanishing gradients when training SPENs defined over the convex relaxation of discrete prediction problems. Sec.~\ref{sec:fast} trains energies such that gradient-based minimization is fast. Sec.~\ref{sec:eff-impl} reduces SPENs' computation and memory overhead. Finally, Sec.~\ref{sec:rec-archs} provides practical recommendations for specific architectures, parameter tying schemes, and pretraining methods that reduce overfitting and improve efficiency.

We demonstrate the effectiveness of our SPEN training methods on two diverse tasks. We first consider depth image denoising on the 7-Scenes dataset~\citep{newcombe2011kinectfusion}, where we employ deep convolutional networks as priors over images. This provides a significant performance improvement, from 36.3 to 40.4 PSNR, over the recent work of~\citep{wang2016proximal}, which unrolls more sophisticated optimization than us, but uses a simpler image prior.  After that, we apply SPENs to semantic role labeling (SRL) on the CoNLL-2005 dataset~\citep{carreras2005introduction}. The task is challenging for SPENs because the output is discrete, sparse, and subject to rigid non-local constraints. We show how to formulate SRL as a SPEN problem and demonstrate performance improvements over strong baselines that use deep features, but sufficiently simple energy functions that the constraints can be enforced using dynamic programming.

Despite substantial differences between the two applications, learning and prediction for all models is performed using the same gradient-based prediction and end-to-end learning code. This black-box interaction with the model provides many opportunities for further use of SPENs.

\section{Structured Prediction Energy Networks}
A SPEN is defined as an instance of Eq.~\eqref{eq:spen} where the energy is given by a deep neural network that provides a subroutine for efficiently evaluating $\frac{d}{d \y} E_{\x}(\y)$~\citep{belanger2016structured}. Differentiability necessitates that the energy is defined on continuous inputs. Going forward, $\y$ will always be continuous. Prediction is performed by gradient-based optimization with respect to $\y$. 

This section first motivates the SPENs employed in this paper, by contrasting them with alternative energy-based approaches to structured prediction. Then, we present two families of methods for training energy-based structured prediction models that have been explored in prior work. 


\subsection{Black-Box vs. Factorized Energy Functions}
\label{sec:fac-energy}
The definition of SPENs above is extremely general and includes many existing modeling techniques. However, both this paper and~\citet{belanger2016structured} depart from most prior work by employing monolithic energy functions that only provide forward and back-propagation. 

This contrasts with the two principal families of energy-based models in the literature, where the tractability of (approximate) energy minimization depends crucially on the factorization structure of the energy. 
First, \textit{factor graphs} decompose the energy into a sum of functions defined over small sets of subcomponents of $\y$~\citep{kschischang2001factor}.
This structure provides opportunities for energy minimization using message passing, MCMC, or combinatorial solvers. Second, \textit{autoregressive models}, such as recurrent neural networks (RNNs) assume an ordering on the components of $\y$ such that the energy for component $\y_i$ only depends on its predecessors. Approximate energy minimization can be performed using search in the space of prefixes of $\y$ using beam search or greedy search. See, for example,~\citet{sutskever2014sequence}.


By not relying on any such factorization when choosing learning and prediction algorithms for SPENs, we can consider much broader families of deep energy functions.  We do not specify the interaction structure in advance, but instead learn it automatically by fitting a deep network. This can capture sophisticated global interactions among components of $\y$ that are difficult to represent using a factorized energy. Of course, the downside of such SPENs is that they provide few guarantees, particularly when employing non-convex energies. Furthermore, for problems with hard constraints on outputs, the ability to do effective constrained optimization may have depended crucially on certain factorization structure. 

\subsection{Learning as Conditional Density Estimation}
One method for estimating the parameters of an energy-based model $E_{\x}(\y)$ is to maximize the conditional likelihood of $\y$:
\begin{align}
\Prob(\y | \x) \propto \exp\left(-E_{\x}(\y)\right).
\end{align}
Unfortunately, computing the likelihood requires the distribution's normalizing constant, which is intractable for black-box energies with no available factorization structure. In~\textit{contrastive backprop}, this is circumvented by performing contrastive divergence training, with Hamiltonian Monte Carlo sampling from the energy surface~\citep{mnih2005learning,hinton2006unsupervised,ngiam2011learning}. Recently,~\citet{zhai2016deep} trained energy-based density models for anomaly detection by exploiting the connections between denosing autoencoders, energy-based models, and score matching~\citep{vincent2011connection}. 


\subsection{Learning with Exact Energy Minimization}
\label{sec:learning-exact}
Let $\Delta(\hat{\y},\y^*)$ be a non-negative task-specific cost function for comparing $\hat{\y}$ and the ground truth $\y^*$.  ~\citet{belanger2016structured} employ a structured SVM (SSVM) loss ~\citep{koller2004max,tsochantaridis2004support}:
\begin{equation}
\sum_{\{\x_i,\y_i\}} \max_{\y} \left[\Delta(\y,\y_i)- E_{\x_i}(\y) + E_{\x_i}(\y_i)\right]_+, \label{eq:ssvm}
\end{equation}
where $\left[\cdot\right]_+ = \max(0,\cdot)$. Each step of minimizing Eq.~\eqref{eq:ssvm} by subgradient descent requires \textit{loss-augmented inference}: 
\begin{equation}
\min_{\y} \left(-\Delta(\y,\y_i) + E_{\x_i}(\y)\right). \label{eq:loss-aug}
\end{equation}
For differentiable $\Delta(\y,\y_i)$,  a local optimum of Eq.~\eqref{eq:loss-aug} can obtained using first-order methods.

Solving Eq.~\eqref{eq:loss-aug} probes the model for margin violations. If none exist, the gradient of the loss with respect to the parameters is zero. Therefore, SSVM performance does not degrade gracefully with optimization errors in the inner prediction problem, since inexact energy minimization may fail to discover margin violations that exist. Performance can be recovered if Eq.~\eqref{eq:loss-aug} returns a lower bound, eg. by solving an LP relaxation~\citep{finley2008training}. However, this is not possible in general. In Sec.~\ref{sec:denoise-results} we compare the image denoising performance of SSVM learning vs. this paper's end-to-end method. Overall, we have found SSVM learning to be unstable and difficult to tune for non-convex energies in applications more complex than the multi-label classification experiments of~\citet{belanger2016structured}.

The~\textit{implicit function theorem} offers an alternative framework for training energy-based predictors~\citep{foo2008efficient,samuel2009learning}. See~\citet{domke2012generic} for an overview. While a naive implementation requires inverting Hessians, one can solve the product of an inverse Hessian and a vector using conjugate gradients, which can leverage the techniques discussed in Sec.~\ref{sec:unrolled} as a subroutine. To perform reliably, the method unfortunately requires exact energy minimization and many conjugate gradient iterations. 

Overall, both of these learning algorithms only update the energy function in the neighborhoods of the ground truth and the predictions of the current model. On the other hand, it may be advantageous to shape the entire energy surface such that is exhibits certain properties, e.g., gradient descent converges quickly when initialized well (Sec.~\ref{sec:fast}). Therefore, these methods may be undesirable even for problems where exact energy minimization is tractable.

For non-convex $E_{\x}(\y)$, gradient-based prediction will only find a local optimum. \citet{amos2016input} present \textit{input-convex neural networks} (ICNNs), which employ an easy-to-implement method for constraining the parameters of a SPEN such that the energy is convex with respect to $\y$, but perhaps non-convex with respect to the parameters. One simply uses convex, non-decreasing non-linearities and only non-negative parameters in any part of the computation graph downstream from $\y$. Here, prediction will return the global optimum, but convexity, especially when achieved this way, may impose a strong restriction on the expressivity of the energy. Their construction is a sufficient condition for achieving convexity, but there are convex energies that disobey this property.  Our experiments present results for instances of ICNNs. In general, non-convex SPENS perform better.

\section{Learning with Unrolled Optimization}
\label{sec:unrolled}

The methods of Sec.~\ref{sec:learning-exact} are unreliable with non-convex energies because we cannot simply use the output of inexact energy minimization as a drop-in replacement for the exact minimizer. Instead, a collection of prior work has performed end-to-end learning of gradient-based predictors~\citep{gregor2010learning,domke2012generic,MacDuvAda2015hyper,andrychowicz2016learning,wang2016proximal,metz2016unrolled,greff2016highway}. Rather than reasoning about the energy minimum as an abstract quantity, the authors pose a specific gradient-based algorithm for approximate energy minimization and optimize its empirical performance using back-propagation. This is a form of~\textit{direct risk minimization}~\citep{tappen2007learning,stoyanov2011empirical,domke2013learning}. 

Consider simple gradient descent: 
\begin{equation}
\y_T = \y_0 - \sum_{t = 1}^T \eta_t \frac{d}{d \y} E_\x(\y_t). \label{eq:gd1}
\end{equation}
To learn the energy function end-to-end, we can back-propagate through the unrolled optimization Eq.~\eqref{eq:gd1} for fixed $T$. 
With this, it can be rendered API-equivalent to a feed-forward network that takes $\x$ as input and returns a prediction for $\y$, and can thus be trained using standard methods. Furthermore, certain hyperparameters, such as the learning rates $\eta_t$, are trainable~\citep{domke2012generic}. 

This backpropagation requires non-standard interaction with a neural-network library because Eq.~\eqref{eq:gd1} computes gradients in the forward pass, and thus it must compute second order terms in the backwards pass. We can save space and computation by avoiding instantiating Hessian terms and instead directly computing Hessian-vector products. These can be achieved three ways. First, the method of~\citet{pearlmutter1994fast} is exact, but requires non-trivial code modifications. Second, some libraries construct computation graphs for gradients that are themselves differentiable. Third, we can employ finite-differences~\citep{domke2012generic}. 

It is clear that Eq.~\eqref{eq:gd1} can be naturally extended to certain alternative optimization methods, such as gradient descent with momentum, or L-BFGS~\citep{liu1989limited,domke2012generic}. These require an additional state vector $\textbf{h}_t$ that is evolved along with $\textbf{y}_t$ across iterations. \citet{andrychowicz2016learning} unroll gradient-descent, but employ a learned non-linear RNN to perform per-coordinate updates to $\y$. End-to-end learning is also applicable to special-case energy minimization algorithms for graphical models, such as mean-field inference and belief propagation~\citep{domke2013learning,chen2014semantic,tompson2014joint,li2014,hershey2014deep,zheng2015conditional}.

\section{End-to-End Learning for SPENs}
\label{sec:e2e-in-practice}
We now present details for applying the methods of the previous section to SPENs. We first describe considerations for learning SPENs defined for the convex relaxation of discrete labeling problems. Then, we describe how to encourage our models to optimize quickly in practice. Finally, we present methods for improving the speed and memory overhead of SPEN implementations.

Our experiments unroll either Eq.~\eqref{eq:gd1} or an analogous version implementing gradient descent with momentum.   We compute Hessian-vector products using the finite-difference method of~\citep{domke2012generic}, which allows black-box interaction with the energy. 

We avoid the RNN-based approach of~\citet{andrychowicz2016learning} because it diminishes the semantics of the energy, as the interaction between the optimizer and gradients of the energy is complicated. In recent work,~\citet{2017arXiv170304363G} propose an alternative learning method that fits the energy function such that $E_\x(\cdot) \approx -\Delta(\cdot,\y^*)$, where $\Delta$ is defined as in Sec.~\ref{sec:learning-exact}. This is an interesting direction for future research, as it allows for non-differentiable $\Delta$. The advantage of end-to-end learning, however, is that it provides a energy function that is precisely tuned for a particular test-time energy minimization procedure.

\subsection{End-to-End Learning for Discrete Problems}
\label{sec:simplex}

To apply SPENs to a discrete structured prediction problem, we relax to a constrained continuous problem, apply SPEN prediction, and then round to a discrete output. For example, for tagging each pixel of a $h \times w$ image with a binary label, we would relax from $\{0,1\}^{w \times h}$ to $[0,1]^{w \times h}$, and if the pixels can take on one of $D$ values, we would relax from $\y \in \{0,\ldots, D\}^{w \times h}$ to $\Delta_D^{w \times h}$, where $\Delta_D$ is the probability simplex on $D$ elements. 

While this rounding introduces poorly-understood sources of error, it has worked well for non-convex energy-based prediction in multi-label classification~\citep{belanger2016structured}, sequence tagging~\citep{vilnis2015bethe}, and translation \citep{DBLP:journals/corr/HoangHC17}.

Both $[0,1]^{w \times h}$ and $\Delta_D^{w \times h}$ are Cartesian products of probability simplices, and it is easy to adopt existing methods for projected gradient optimization over the simplex.

First, it is natural to apply Euclidean projected gradient descent. Over $[0,1]^{w \times h}$, we have:
\begin{equation}
\y_{t+1} = \text{Clip}_{0,1}\left[\y_t  - \eta_t \nabla E_{\x}(\y_t)\right],
\end{equation}
This is unusable for end-to-end learning, however, since back-propagation through the projection will yield 0 gradients whenever $\y_t  - \eta_t \nabla E_{\x}(\y_t) \notin [0,1]$. This is similarly problematic for projection onto $\Delta_D^{w \times h}$~\citep{duchi2008efficient}.

Alternatively, we can apply entropic mirror descent, ie. projected gradient with distance measured by KL divergence~\citep{beck2003mirror}. For $\y \in \Delta_D^{w \times h}$, we have:
\begin{equation}
\y_{t+1} = \text{SoftMax}\left(\log(\y_t)  - \eta_t \nabla E_{\x}(\y_t)\right) \label{eq:emd}
\end{equation}
This is suitable for end-to-end learning, but the updates are similar to an RNN with sigmoid non-linearities, which is vulnerable to vanishing gradients~\citep{bengio1994learning}. 

Instead, we have found it useful to avoid constrained optimization entirely, by optimizing un-normalized logits $\mathbf{l}_t$, with $\y_t = \text{SoftMax}(\mathbf{l}_t)$:
\begin{equation}
\mathbf{l}_{t+1} = \mathbf{l}_t - \eta_t\nabla E_{\x}\left(\text{SoftMax}(\mathbf{l}_t)\right). \label{eq:logit}
\end{equation}
Here, the updates to $\mathbf{l}_t$ are additive, and thus will be less susceptible to vanishing gradients~\citep{hochreiter1997long,srivastava2015training,he2016deep}. 

Finally,~\citet{amos2016input} present the \textit{bundle entropy method} for convex optimization with simplex constraints, along with a method for differentiating the output of the optimizer. End-to-end learning for Eq.~\eqref{eq:gd} can be performed using generic learning software, since the unrolled optimization obeys the API of a feed-forward predictor, but unfortunately this is not true for their method. Future work should consider their method, however, as it performs very rapid energy minimization.


\subsection{Learning to Optimize Quickly}     
\label{sec:fast}

We next enumerate methods for learning a model such that gradient-based energy minimization converges to high-quality $\y$ quickly. When using such methods, we have found it important to maintain the same optimization configuration, such as $T$, at both train and test time. 

First, we can encourage rapid optimization by defining our loss function as a sum of losses on every iterate $\y_t$, rather than only on the final one. Let $\ell(\y_t,\y^*)$ be a differentiable loss between an iterate and the ground truth. We employ
\begin{equation}
L = \frac{1}{T}\sum_{t=1}^Tw_t\ell(\y_t,\y^*), \label{eq:avg-loss}
\end{equation}
where $w_t$ is a non-negative weight. This encourages the model to achieve high-quality predictions early. It has the additional benefit that it reduces vanishing gradients, since a learning signal is introduced at every timestep. Our experiments use $w_t = \frac{1}{T - t + 1}$. 

Second, for the simplex-constrained problems of Sec.~\ref{sec:simplex}, we smooth the energy with an entropy term $\sum_i H(\y_i)$. This introduces extra strong convexity, which helps improve convergence. It also strengthens the parallel between SPEN prediction and marginal inference in a Markov random field, where the inference objective is expected energy plus entropy~\citep[p.~385]{koller2009probabilistic}. 

Third, we can set $T$ to a small value. Of course, this guarantees that optimization converges quickly on the train data. Here, we lose the contract that Eq.~\eqref{eq:gd} is even performing energy minimization, since it hasn't converged, but this may be acceptable if predictions are accurate. For example, some experiments achieve good performance with $T = 3$.

In future work, it may be fruitful to directly penalize convergence criteria, such as $\lVert \y_{t} - \y_{t-1}\rVert$ and $ \lVert \frac{d}{d\y_t} E_\x(\y_t) \rVert$.


\subsection{Efficient Implementation}
\label{sec:eff-impl}
Since we can explicitly encourage our model to converge quickly, it is important to exploit fast convergence at train time. Eq.~\eqref{eq:gd} is unrolled for a fixed $T$. However, if optimization converges at $T_0 < T$, it suffices to start back-propagation at $T_0$, since the updates to $\y_t$ for $t > T_0$ are the identity. Therefore, we unroll for a fixed number of iterations $T$, but iterate only until convergence is detected. 

To support back-propagation, a naive implementation of Eq.~\eqref{eq:gd} would require $T$ clones of the energy (with tied parameters). We reduce memory overhead by checkpointing the inputs and outputs of the energy, but discarding its internal state. This allows us to use a single copy of the energy, but requires recomputing forward evaluations at specific $\y_t$ during the backwards pass. To save additional memory, we could have reconstructed the $\y_t$ on-the-fly either by reversing the dynamics of the energy minimization method~\citep{domke2013learning,MacDuvAda2015hyper} or by performing a small amount of extra forward-propagation~\citep{rey2000exact,lewis2003debugging}. 


\section{Recommended SPEN Architectures for End-to-End Learning}
\label{sec:rec-archs}
 To train SPENs end-to-end, we write Eq.~\eqref{eq:gd1} as:
\begin{equation}
\y_T = \text{Init}(F(\x)) - \sum_{t = 1}^T \eta_t \frac{d}{d \y} E(\y_t \; ; \; F(\x)). \label{eq:gd}
\end{equation}
Here, $\text{Init}(\cdot)$ is a differentiable procedure for predicting an initial iterate $\y_0$. 
Following~\citet{belanger2016structured}, we also employ $E_{\x}(\y) = E(\y \; ; \; F(x))$,  where the dependence of $E_{\x}(\y)$ on $\x$ comes by way of a parametrized feature function $F(\x)$. This is useful because test-time prediction can avoid back-propagation in $F(x)$. 

We have found it useful in practice to employ an energy that splits into global and local terms:
\begin{equation}
E(\y \; ;F(\x) ) = E^g(\y \; ;F(\x))  + \sum_i  E_i^l(\y_i \; ;F(\x)).
\end{equation}
Here, $i$ indexes the components of $\y$ and $E^g(\y \; ;F(\x))$ is an arbitrary global energy function.  The modeling benefits of the local terms are similar to the benefits of using local factors in popular factor graph models. We also can use the local terms to provide an implementation of $\text{Init}(\cdot)$.
%
%

We pretrain $F(x)$ by training the feed-forward predictor $\text{Init}(F(x))$. We also stabilize learning by first clamping the local terms for a few epochs while updating $E^g(\y \; ;F(\x))$.

To back-propagate through Eq.~\eqref{eq:gd}, the energy function must be at least twice differentiable with respect to $\y$. Therefore, we can't use non-linearities with discontinuous gradients. Instead of ReLUs, we use a SoftPlus with a reasonably high temperature. Note that $F(\x)$ and $\text{Init}(\cdot)$ can be arbitrary networks that are sub-differentiable with respect to their parameters.

\section{Experiments}
We evaluate SPENs on image denoising and semantic role labeling (SRL) tasks. Image denoising is an important benchmark for SPENs, since the task appears in many prior works employing end-to-end learning. SRL is useful for evaluating SPENs' suitability for challenging combinatorial problems, since the outputs are subject to rigid, non-local constraints.  For both, we provide controlled experiments that isolate the impact of various SPEN design decisions, such as the optimization method that is unrolled and the expressivity of the energy function.

 In these applications, we employ specific architectures based on our prior knowledge about the problem domain. This capability is crucial for introducing the necessary inductive bias to fbe able to fit SPENs on limited datasets. Overall, black-box prediction and learning methods for SPENs are useful because we can select architectures based on their suitability for the data, not whether they support model-specific algorithms. 


\subsection{Image Denoising}
\label{sec:denoising}


Let $\x \in [0,1]^{w \times h}$ be an observed grayscale image. We assume that it is a noisy realization of a latent clean image $\y \in [0,1]^{w \times h}$, which we estimate using MAP inference. Consider a Gaussian noise model with variance $\sigma^2$ and a prior $\Prob(\y)$. The associated energy function is: 
\begin{equation}
\lVert \y - \x \rVert_2^2 - 2\sigma^{2}\log \Prob(\y). \label{eq:MAP}
\end{equation}
Here, the feature network is the identity. The first term is the local energy network and the second, which does not depend on $\x$, is the global energy network. 

There are three general families for the prior. First, it can be hard-coded. Second, it can be learned by approximate density estimation. Third, given a collection of $\{\x,\y\}$ pairs, we can perform supervised learning, where the prior's parameters are discriminatively trained such that the output of a particular algorithm for minimizing Eq.~\eqref{eq:MAP} is high-quality. End-to-end learning has proven to be highly successful for the third approach~\citep{tappen2007learning,barbu2009training,schmidt2010generative,sun2011learning,domke2012generic,wang2016proximal}, and thus it is important to evaluate the methods of this paper on the task. 

\subsubsection{Image Priors}
Much of the existing work on end-to-end training for denoising considers some form of a field-of-experts (FOE) prior~\citep{roth2005fields}. We consider an $\ell_1$ version, which assigns high probability to images with sparse activations from $K$ learned filters:
\begin{equation}
\Prob(\y) \propto \exp\left(-\sum_k  \lVert(\mathbf{f}_k \ast \y)  \rVert_1\right). \label{eq:foe}
\end{equation}
\citet{wang2016proximal} perform end-to-end learning for Eq.~\eqref{eq:foe}, by unrolling proximal gradient methods that analytically handle the non-differentiable $\ell_1$ term. 

This paper assumes we only have black-box interaction with the energy. In response, we alter Eq.~\eqref{eq:foe} such that it is twice differentiable, so that we can unroll generic first-order optimization methods. We approximate Eq.~\eqref{eq:foe} by leveraging a SoftPlus with temperature 25, replacing $|\cdot|$ by: 
\begin{equation}
\text{SoftAbs}(\y) = 0.5 \; \text{SoftPlus}(\y) + 0.5\;\text{SoftPlus}(-\y). \label{eq:softabs}
\end{equation}

The principal advantage of learning algorithms that are not hand-crafted to the problem structure is that they provide the opportunity to employ more expressive energies. In response, we also consider a deeper prior, given by:
\begin{equation}
\Prob(\y) \propto \exp\left(-\text{DNN}(\y) \right). \label{eq:dnn}
\end{equation}
Here, $\text{DNN}(\y)$ is a general deep convolutional network that takes an image and returns a number. The architecture in our experiments consists of a $7 \times 7 \times 32$  convolution, a SoftPlus, another $7 \times 7 \times 32$ convolution, a SoftPlus, a $1 \times 1 \times 1$ convolution, and finally spatial average pooling. The method of~\citet{wang2016proximal} cannot handle this prior. 

\subsubsection{Experimental Setup}
We evaluate on the 7-Scenes dataset~\citep{newcombe2011kinectfusion}, where we seek to denoise depth measurements from a Kinect sensor. Our data processing and hyperparameters are designed to replicate the setup of~\citet{wang2016proximal}, who demonstrate state-of-the art results for energy-minimization-based denoising on the dataset. We train using random 96 $\times$ 128 crops from 200 images of the same scene and report PSNR (higher is better) for 5500 images from different scenes. We treat $\sigma^2$ as a trainable parameter and minimize the mean-squared-error of $\y$.

\subsubsection{Results and Discussion}
\label{sec:denoise-results}
\begin{figure}[tb]
\centering
\begin{tabular}{cc}
  \includegraphics[width=0.4\columnwidth]{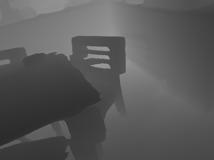} &
  \includegraphics[width=0.4\columnwidth]{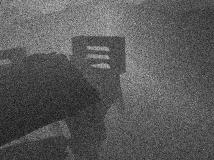} \\
{\small Ground Truth } & {\small Noisy Input}  \\
  \includegraphics[width=0.4\columnwidth]{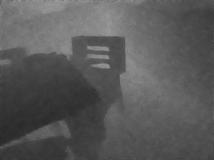} &
  \includegraphics[width=0.4\columnwidth]{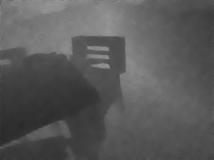} \\
{\small FOE-20 } & {\small FOE-20+}  \\
\includegraphics[width=0.4\columnwidth]{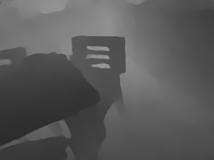} & 
\includegraphics[width=0.4\columnwidth]{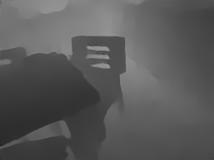} \\
{\small DeepPrior-20 } & {\small DeepPrior-3}  \\
\end{tabular}
\caption{Example Denoising Outputs}
\label{fig:denoise}
\end{figure}

Example outputs are given in Figure~\ref{fig:denoise} and Table~\ref{table:denoise} compares PSNR. \textbf{BM3D} is a widely-used non-parametric method~\citep{dabov2007image}. FilterForest (\textbf{FF}) adaptively selects denoising filters for each location~\citep{ryan2014filter}. ProximalNet (\textbf{PN}) is the system of~\citet{wang2016proximal}. \textbf{FOE-20} is an attempt to replicate \textbf{PN} using end-to-end SPEN learning. We unroll 20 steps of gradient descent with momentum 0.75 and use the modification in Eq.~\eqref{eq:softabs}. Note it performs similarly to \textbf{PN}, which unrolls 5 iterations of sophisticated optimization. Note that we can obtain 37.0 PSNR using a feed-forward convnet with a similar architecture to our DeepPrior, but without spatial pooling.

The next set of results consider improved instances of the FOE model. First, \textbf{FOE-20+} is identical to \textbf{FOE-20}, except that it employs the average loss Eq.~\eqref{eq:avg-loss}, uses a momentum constant of 0.25, and treats the learning rates $\eta_t$ as trainable parameters. We find that this results in both better performance and faster convergence. Of course, we could achieve fast convergence by simply setting $T$ to be small. In response, we consider \textbf{FOE-3}. This only unrolls for $T = 3$ iterations and obtains superior performance. 

The final three results are with the DNN prior Eq.~\eqref{eq:dnn}. \textbf{DP-20} unrolls 20 steps of gradient descent with a momentum constant of 0.25. The gain in performance is substantial, especially considering that a PSNR of 30 can be obtained with elementary signal processing. Similar to \textbf{FOE-3} vs. \textbf{FOE-20+}, we experience a modest performance gain using \textbf{DP-3}, which only unrolls for 3 gradient steps but is otherwise identical. 

Finally, the \textbf{FOE-SSVM} and \textbf{DP-SSVM} configurations use SSVM training. We find that  \textbf{FOE-SSVM} performs competitively with the other FOE configurations. This is not surprising, since the FOE prior is convex. However, fitting the DeepPrior with an SSVM is inferior to using end-to-end learning.  The performance is very sensitive to the energy minimization hyperparameters. 

In these experiments, it is superior to only unroll for a few iterations for end-to-end learning. One possible reason is that a shallow unrolled architecture is easier to train. Truncated optimization with respect to $\y$ may also provide an interesting prior over outputs~\citep{duvenaud2016early}. It is also observed in~\citet{wang2014efficient} that better energy minimization for FOE models may not improve PSNR. Often unrolling for 20 iterations results in over-smoothed outputs.

We are unable achieve reasonable performance with an ICNN~\citep{amos2016input}, which restricts all of the parameters of the convolutions to be positive. Unfortunately, this hinders the ability of the filters in the prior to act as edge detectors or encourage local smoothness. Both of these are important for high-quality denoising. Note that the $\ell_1$ FOE is convex, even without the restrictive ICNN constraint.


\begin{table}[tb!]
\begin{footnotesize}
\begin{center}
\begin{tabular}{|c|c|c|c| c |}
\hline
\textbf{BM3D} & \textbf{FF} & \textbf{PN} & \textbf{FOE-20} & \textbf{FOE-SSVM}\\
\hline
35.46 & 35.63 & 36.31 & 36.41 & 37.7\\
\hline
\hline
\textbf{FOE-20+} & \textbf{FOE-3} & \textbf{DP-20} & \textbf{DP-3}  & \textbf{DP-SSVM} \\
\hline
37.34 & 37.62 & 40.3 & \textbf{40.4}  & 38.7\\
\hline
\end{tabular}
\end{center}
\caption{Denoising Results (PSNR)}
\label{table:denoise}
\end{footnotesize}
\end{table}

\subsection{Semantic Role Labeling}

Semantic role labeling (SRL) predicts the semantic structure of predicates and arguments in sentences~\citep{gildea2002automatic}. For example, in the sentence ``I want to buy a car," the verbs ``want" and ``buy" are two predicates, and ``I" is an argument that refers to the wanter and buyer, ``to buy a car" is the thing wanted, and ``a car" is the thing bought. Given predicates, we seek to identify arguments and their semantic roles in relation to each predicate. Formally, given a set of predicates $\mathbf{p}$ in a sentence $\x$ and a set of candidate argument spans $\ab$, we assign a discrete semantic role $\rb$ to each pair of predicate and argument, where $\rb$ can be either a pre-defined role label or an empty label. We evaluate SRL instead of, for example, noun-phrase chunking~\citep{lacoste2012block}, since it is a more challenging task, 
where the outputs are subject to substantially more complex non-local constraints.

Existing work imposes hard constraints on $\rb$, such as excluding overlapping arguments and repeated core roles during prediction. The objective is to minimize the energy:
\begin{equation}
\min_{\rb} E(\rb \; ; \x,\pb,\ab) \; \text{s.t.}\;\;\rb \in \mathcal{Q}(\x,\pb,\ab), \label{eq:srl-eb}
\end{equation}
where $\mathcal{Q}(\x,\pb,\ab)$ is set of feasible joint role assignments. This constrained optimization problem can be solved using integer linear programming (ILP)~\citep{punyakanok2008importance} or its relaxations~\citep{das2012exact}. These methods rely on the output of local classifiers that are unaware of structural constraints during training. More recently,~\citet{tackstrom2015efficient} account for the constraint structure using dynamic programming at train time.~\citet{fitzgerald2015semantic} extend this using neural network features and show improved results. 

\subsubsection{Data and Preprocessing and Baselines}
We consider the CoNLL 2005 shared task data~\citep{carreras2005introduction}, with standard data splits and official evaluation scripts. We apply similar preprocessing as~\citet{tackstrom2015efficient}. This includes part-of-speech tagging, dependency parsing, and using
the parse to generate candidate arguments. 

Our baseline is an arc-factored model for the conditional probability of the predicate-argument arc labels:
\begin{equation}
\Prob(\rb|\x,\pb,\ab) = \Pi_i \Prob(\rb_i |\x,\pb,\ab). \label{eq:arc-factored}
\end{equation}
where $\Prob(r_i |\x,\pb,\ab)\propto \exp\big(g(\rb_i, \x,\pb,\ab) \big)$. Here, each conditional distribution is given by a multiclass logistic regression model. See Appendix~\ref{app:srl-feat-arch} for details of the architecture and training procedure for our baseline.



When using the negative log of Eq.~\eqref{eq:arc-factored} as an energy in Eq.~\eqref{eq:srl-eb}, there are variety of methods for finding a near-optimal $\rb \in \mathcal{Q}(\x,\pb,\ab)$. First, we can employ simple heuristics for locally resolving constraint violation. The \textbf{Local + H} system uses Eq.~\eqref{eq:arc-factored} and these. We can instead use the AD$^3$ message passing algorithm~\citep{martins2011augmented} to solve the LP relaxation of this constrained problem. We use \textbf{Local + AD}$^3$ to refer to this system. Since the LP solution may not be integral, we post-process the AD$^3$ output using the same heuristics as \textbf{Local + H}. 


\subsubsection{SPEN Model}

The SPEN performs continuous optimization over the relaxed set $\y_i \in \Delta_A$ for each discrete label $\rb_i$, where $A$ is the number of possible roles. The preprocessing generates sparse predicate-argument candidates, but we optimize over the complete bipartite graph between predicates and arguments to support vectorization. We have $\y \in \Delta_A^{n \times m}$, where $n$ and $m$ are the max number of predicates and arguments. Invalid arcs are constrained to the empty label.

We employ a pretrained version of Eq.~\eqref{eq:arc-factored} to provide the local energy term of a SPEN. This is augmented with global terms that couple the outputs together. See Appendix~\ref{sec:srl-arch} for details of the architecture we use. It has terms, for example, that apply a deep network to the feature representations of all of the arcs selected for a given predicate. 

As with~\citet{tackstrom2015efficient}, we seek to account for constraints $\mathcal{Q}(\x,\pb,\ab)$ during both inference and learning, rather than only imposing them via post-processing. Therefore, we include additional energy terms that encode membership in $\mathcal{Q}(\x,\pb,\ab)$ as twice-differentiable soft constraints that can be applied to $\y$. All of the constraints in $\mathcal{Q}(\x,\pb,\ab)$ express that certain arcs cannot co-occur. For example, two arguments cannot attach to the same predicate if the arguments correspond to spans of tokens that overlap. Consider general binary variables $a$ and $b$ with corresponding relaxations $\bar{a},\bar{b} \in [0,1]$. We convert the constraint $\lnot (a \land b)$ into an energy function $\alpha\text{SoftPlus}(\bar{a} + \bar{b} - 1)$, where $\alpha$ is a learned parameter. 

We consider the \textbf{SPEN + H} and \textbf{SPEN + AD}$^3$ configurations, which employ heuristics or AD$^3$ to enforce the output constraints. Rather than applying these methods to the probabilities from Eq.~\eqref{eq:arc-factored}, we use the soft prediction output by energy minimization. 

\begin{table}
\begin{footnotesize}
\begin{center}
\begin{tabular}{|c|c|c|c|}
\hline
& Dev  & Test  & Test\\
Model & (WSJ) & (WSJ) & (Brown) \\
\hline
{\bf Local + H} & 78.0 & 79.7 & 69.7\\
\hline
{\bf Local + AD}$^3$ & 78.2 & 80.0 & 69.9\\
\hline
{\bf SPEN + H} & 79.0 & 80.7 & 69.3\\
\hline
{\bf SPEN + AD}$^3$ & \textbf{79.0} & \textbf{80.7} & 69.4\\
\hline
\hline
T{\"a}ckstr{\"o}m (Local) & 77.9 & 79.3 & 70.2\\
\hline
T{\"a}ckstr{\"o}m (Structured) & 78.6 & 79.9 & \textbf{71.3}\\
\hline
FitzGerald (Local) & 78.4 & 79.4 & 70.9\\
\hline
FitzGerald (Structured) & 78.3 & 79.4 & 71.2\\
\hline
\end{tabular}
\end{center}
\caption{SRL Results (F1)}
\label{table:srl}
\end{footnotesize}
\end{table}


\subsubsection{Results and Discussion}
Table~\ref{table:srl} contains results on the CoNLL 2005 WSJ dev and test sets and the Brown test set. We compare the \textbf{SPEN} and \textbf{Local} systems with the best non-ensemble systems of~\citet{tackstrom2015efficient} and~\citet{fitzgerald2015semantic}, which have similar overall setups as us for feature extraction and for the parametrization of the local energy terms. For these,  `Local' fits Eq.~\eqref{eq:arc-factored} without regard for the output constraints, whereas `Structured' explicitly considers them during training. Note that~\citet{zhou2015end} obtain slightly better performance with alternative RNN methods. We were unable to outputerform the \textbf{Local} systems using a \textbf{SPEN} system trained with an SSVM loss.

We select our SPEN configuration by maximizing performance of \textbf{SPEN + AD}$^3$ on the dev data. Our best system unrolls for 10 iterations, trains per-iteration learning rates, uses no momentum, and unrolls Eq.~\eqref{eq:logit}.  Overall, \textbf{SPEN + AD}$^3$ performs the best of all systems on the WSJ test data. We expect our diminished performance on the Brown test set is due to overfitting. The Brown set is not from the same source as the train, dev, and test WSJ data. SPENs are more susceptible to overfitting because the expressive global term introduces many parameters. 

Note that \textbf{SPEN + AD}$^3$ and \textbf{SPEN + H} performs identically, whereas \textbf{LOCAL + AD}$^3$ and \textbf{LOCAL + H} do not. This is because our learned global energy encourages constraint satisfaction during gradient-based optimization of $\y$. Using the method of~\citet{amos2016input} for restricting the energy to be convex wrt $\y$, we obtain 80.3 on the test set. 


\section{Conclusion and Future Work}
SPENs are a flexible, expressive framework for structured prediction, but training them can be challenging. This paper provides a new end-to-end training method that enables high performance on considerably more complex tasks than those of~\citet{belanger2016structured}.  We unroll an approximate energy minimization algorithm into a differentiable computation graph that is trainable by gradient descent. The approach is user-friendly in practice because it returns not just an energy function but also a test-time prediction procedure that has been tailored for it. 

In the future, it may be useful to employ more sophisticated unrolled optimizers, perhaps where the optimizer's hyperparameters are a learned function of $\x$, and to perform iterative optimization in a learned feature space, rather than output space. Finally, we could model gradient-based prediction as a sequential decision making problem and train the energy using value-based reinforcement learning. 

\section*{Acknowledgments}
Many thanks to Justin Domke, Tim Vieiria, Luke Vilnis, and Shenlong Wang for helpful discussions. The first and third authors were supported in part by the Center for Intelligent Information Retrieval and in part by DARPA under agreement number FA8750-13-2-0020.  The second author was supported in part by DARPA under contract number FA8750-13-2-0005. The U.S. Government is authorized to reproduce and distribute reprints for Governmental purposes notwithstanding any copyright notation thereon. Any opinions, findings and conclusions or recommendations expressed in this material are those of the authors and do not necessarily reflect those of the sponsor.

\bibliography{sources}
\bibliographystyle{icml2017}

\clearpage
\appendix

\section{Appendix}
\subsection{General Learning Setup}
The method described in Sections~\ref{sec:unrolled} and~\ref{sec:e2e-in-practice} provides a gradient of the loss with respect to the parameters of the model. To update the parameters, one can use any standard optimization method for neural networks. Our experiments use Adam~\cite{kingma2014adam} with default settings. SPENs are vulnerable to overfitting, as the energy network is often very expressive. We reduce overfitting by performing early stopping, by taking the model that performs best on development data. Often, we have found that early stopping with a model that has a higher capacity energy (e.g., higher-dimensional hidden layers in the energy network) is superior to using a low-capacity energy. 

\subsection{Architectures for SRL Experiments}
\subsubsection{Baseline Arc-Factored Architecture}
\label{app:srl-feat-arch}
Our baseline is an arc-factored model for the conditional probability of the predicate-argument arc labels:
\begin{equation}
\Prob(\rb|\x,\pb,\ab) = \Pi_i \Prob(r_i |\x,\pb,\ab). \label{eq:arc-factored}
\end{equation}
where $\Prob(r_i |\x,\pb,\ab)\propto \exp\big(g(r_i, \x,\pb,\ab) \big)$. Here, each conditional distribution is given by a logistic regression model. We compute $g(r_i,\x,\pb,\ab)$ using a multi-layer perceptron (MLP) similar to~\citet{fitzgerald2015semantic}. Its inputs are discrete features extracted from the argument span and the predicate (including words, pos tags, and syntactic dependents), and the dependency path and distance between the argument and the predicate.  These features are transformed to a 300-dimensional representation linearly, where the embeddings of word types are initialized using newswire embeddings from~\cite{mikolov2013distributed}. We map from 300 dimensions to 250 to 47 (the number of semantic roles in CoNLL) using linear transformations separated by tanh layers. We apply dropout to the embedding layer with rate 0.5 and a standard log loss.

\subsubsection{Global Energy Term for SPEN}
\label{sec:srl-arch}
From the pre-trained model Eq.~\eqref{eq:arc-factored}, we define $\mathbf{f}_r$ as the predicate-argument arc features, We also have predicate features $\mathbf{f}_p$ and argument feature $\mathbf{f}_a$, given by the average word embedding of the token spans. 
The hidden layers of any MLP below are 50-dimensional. Each MLP is two layers, with a SoftPlus in the middle. All parameters are trained discriminatively using end-to-end training.

Let $\y_p \in \Delta_A^m$ be the sub-tensor of $\y$ for a given predicate $p$ and let $\z_p = \sum_k \y_p[:,k] \in [0,1]^m$, where $\z_{p}[a]$ is the total amount of mass assigned to the arc between predicate $p$ and argument $a$, obtained by summing over possible labels. We also define  $\w_p= \sum_k \y_p[k,:]\in \reals_+^A$. This is a length-A vector containing how much total mass of each arc label is assigned to predicate $p$. Finally, define $\mathbf{s}_r = \sum_k \y[:,:,k]$. This is the total mass assigned to arc $r$, obtained by summing over the possible labels that the arc can take on. 

The global energy is defined by the sum of the following terms. The first energy term scores the set of arguments attached to each predicate. It computes a weighted average of the features $\textbf{f}_a$ for the arguments assigned to predicate $p$, with weights given by $\z_p$. It then concatenates this with $\mathbf{f}_p$, and passes the result through a two-layer multi-layer perceptron (MLP) that returns a single number. The total energy is the sum of the MLP output for every predicate. The second energy term scores the labels of the arcs attached to each predicate. We concatenate $\mathbf{f}_p$ with $\w_p$ and pass this through an MLP as above. The third energy term models how many arguments a predicate should take on. For each predicate, we predict how many arguments should attach to it, using a linear function applied to $\mathbf{f}_p$. The energy is set to the squared difference between this and the total mass attached to the predicate under $\y$, which is given by $\sum_k \w_p[k]$. The fourth energy term averages $\w_p$ over all $p$ and applies an MLP to the result. The fifth term computes a weighted average of the arc features $\textbf{f}_r$, with weights given by $\mathbf{s}_r$ and also applies an MLP to the result. The last two terms capture general topical coherence of the prediction. 

\end{document}